\documentclass[letterpaper, 10 pt, conference]{ieeeconf}  

\IEEEoverridecommandlockouts                              

\usepackage{multirow}
\usepackage{hyperref}       
\usepackage{url}            
\usepackage{booktabs}       
\usepackage{amsfonts}       
\usepackage{amsmath}       
\usepackage{graphicx}       
\usepackage{nicefrac}       
\usepackage{microtype}      
\usepackage{xcolor}
\usepackage{caption}
\usepackage{dblfloatfix}
\usepackage{placeins}
\usepackage{threeparttable}

\title{\LARGE \bf
OMNI-PoseX: A Fast Vision Model for 6D Object Pose Estimation \\ in Embodied Tasks
}

\author{Michael Zhang$^{1}$, Wei Ying$^{1}$, Fangwen Chen$^{1}$, Shifeng Bai$^{1}$, Hanwen Kang$^{1*}$%
\thanks{$^{1}$KernalMind Tech Co., Ltd., Shanghai, 201100, China.}
\thanks{$^{*}$Corresponding author}%
}

\begin{document}

\makeatletter
\let\old@maketitle\@maketitle
\renewcommand{\@maketitle}{%
\old@maketitle
\vspace{-0.2cm}
\begin{center}
\captionsetup{type=figure}
\setcounter{figure}{0}
\includegraphics[width=\textwidth]{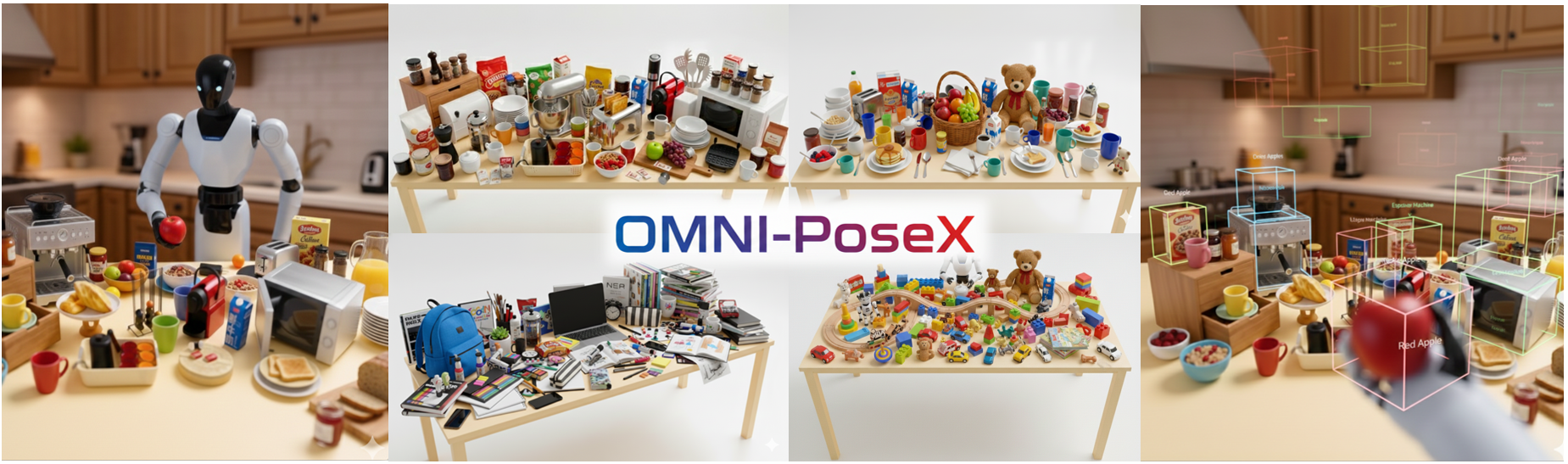}
\captionof{figure}{OMNI-PoseX is a vision foundation model for 6D pose estimation in embodied tasks. It predicts object categories, 3D bounding boxes, and 6D poses. Trained on large-scale open-source datasets, OMNI-PoseX generalizes robustly across objects and embodied scenarios.}
\label{fig:overview}
\end{center}
\vspace{-0.3cm}
}
\makeatother

\maketitle

\begin{abstract}
Accurate 6D object pose estimation is a fundamental capability for embodied agents, yet remains highly challenging in open-world environments. Many existing methods often rely on closed-set assumptions or geometry-agnostic regression schemes, limiting their generalization, stability, and real-time applicability in robotic systems.
We present \textbf{OMNI-PoseX}, a vision foundation model that introduces a novel network architecture unifying open-vocabulary perception with an SO(3)-aware reflected flow matching pose predictor. The architecture decouples object-level understanding from geometry-consistent rotation inference, and employs a lightweight multi-modal fusion strategy that conditions rotation-sensitive geometric features on compact semantic embeddings, enabling efficient and stable 6D pose estimation. To enhance robustness and generalization, the model is trained on large-scale 6D pose datasets, leveraging broad object diversity, viewpoint variation, and scene complexity to build a scalable open-world pose backbone.
Comprehensive evaluations across benchmark pose estimation, ablation studies, zero-shot generalization, and system-level robotic grasping integration demonstrate the effectiveness of OMNI-PoseX. The \textbf{OMNI-PoseX} achieves SOTA pose accuracy and real-time efficiency, while delivering geometrically consistent predictions that enable reliable grasping of diverse, previously unseen objects.

\end{abstract}

\section{INTRODUCTION}
Estimating the 6D pose of objects is a core perceptual capability for embodied agents, underpinning a wide range of downstream tasks such as manipulation\cite{argenziano2024empower}\cite{du2021vision}\cite{tremblay2018deep}, navigation\cite{gu2024conceptgraphs}, and physical interaction\cite{yang2025mk}. In realistic embodied environments, robots must infer object pose under diverse viewpoints, severe occlusions, cluttered scenes, and partial observations, often while interacting with objects whose appearances and geometries differ significantly from those seen during training. These challenges are further amplified in open-world settings, where object categories and instances cannot be exhaustively enumerated in advance, placing stringent demands on generalization and robustness beyond closed-set assumptions.

Despite recent advances, 6D pose estimation in open-world embodied scenarios remains fundamentally challenging. Many existing approaches rely on closed-set object categories\cite{moon2025co}\cite{lunayach2024fsd}\cite{zhang2022rbp}, instance-specific supervision\cite{liu2022catre}\cite{deng2025pos3r}\cite{fu2022category}, or predefined object templates\cite{wen2024foundationpose}\cite{wang2021occlusion}\cite{zakharov2019dpod}\cite{song2020hybridpose}, which limits their applicability to unconstrained environments with unseen objects. Moreover, recent generative formulations that model object orientation through stochastic score estimation on the SO(3) manifold typically require iterative denoising with carefully tuned noise schedules\cite{ikeda2024diffusionnocs}. While effective in improving rotational accuracy, such methods incur substantial computational overhead and often exhibit unstable inference dynamics, making them less suitable for real-time embodied tasks where both efficiency and temporal consistency are critical.

In this work, we introduce \textbf{OMNI-PoseX}, a vision foundation model for open-world 6D pose estimation that centers geometric structure in pose inference; we formulate object orientation prediction as an SO(3)-aware reflected flow matching problem, modeling rotations as geodesic flows in the Lie algebra to ensure valid trajectories and avoid linear interpolation artifacts; this formulation yields a stable, non-stiff velocity field that enables accurate inference with a small, fixed number of integration steps; OMNI-PoseX decouples open-world perception from geometry-aware pose reasoning by combining open-vocabulary visual perception with an SO(3)-aware pose network through multi-modal feature fusion; the model is trained on a large-scale 6D pose dataset integrating real-world and simulated scenes; extensive experiments show that OMNI-PoseX outperforms existing methods on open-world benchmarks, generalizes robustly to unseen objects, and produces temporally stable pose estimates in downstream embodied manipulation tasks.
Our contributions are threefold:
\begin{itemize}
    \item we propose an SO(3)-aware reflected flow matching framework for efficient and stable 6D pose estimation;

    \item we present an open-world vision foundation model that unifies object recognition, 3D bounding box estimation, and geometry-consistent pose inference;

    \item we demonstrate strong zero-shot generalization and robust performance on large-scale benchmarks and downstream embodied manipulation tasks..
\end{itemize}

\section{Related works}
\label{related_works}
6D object pose estimation can be broadly categorized into \emph{instance-level} and \emph{category-level} approaches\cite{peng2022self}\cite{tian2020shape}. Instance-level methods assume access to object CAD models and estimate poses for seen instances, limiting generalization to novel objects. Category-level approaches instead learn category priors from diverse instances and enable pose estimation for unseen objects within known categories without CAD models. Representative works such as NOCS~\cite{wang2019normalized}, SPD~\cite{wang2021category}, SGPA~\cite{chen2021sgpa}, HS-Pose~\cite{zheng2023hs}, and IST-Net~\cite{liu2023net} progressively improve category-level reasoning through coordinate normalization, prior modeling, and feature transformation. Generative methods such as GenPose~\cite{zhang2023generative}\cite{zhang2024omni6dpose} further address symmetry ambiguity by modeling pose distributions.

Despite these advances, two fundamental challenges remain.
First, object orientation lies on the compact Lie group $SO(3)$, whose non-Euclidean geometry and non-trivial topology fundamentally differ from vector spaces \cite{simeonov2023se}. Direct regression of axis-angle, quaternion, or rotation matrices under Euclidean losses implicitly assumes linear structure, which may introduce representation discontinuities, ambiguous gradients, and biased interpolation that deviates from geodesics on the manifold \cite{fang1billion}. While diffusion- or score-based approaches define stochastic processes directly on $SO(3)$ to better respect its geometry, they typically require carefully designed noise schedules and iterative denoising with many integration steps, leading to high computational overhead and limited real-time applicability \cite{yim2023fast}.
Second, although large-scale vision pretraining and open-vocabulary learning have significantly enhanced semantic generalization, most pose estimation pipelines remain tightly coupled to closed-set assumptions \cite{wen2024foundationpose}. Pose modules are often trained jointly with task-specific perception components, restricting scalability across unseen categories and diverse embodied environments. A principled integration of open-world perception with geometry-aware pose modeling remains underexplored.

OMNI-PoseX addresses these challenges within a unified framework. We introduce an \emph{SO(3)-aware reflected flow matching} formulation that learns a geometry-consistent velocity field directly in the Lie algebra, supervising rotational velocities rather than absolute poses. By constructing time-symmetric geodesic flows, the method preserves intrinsic manifold structure while enabling stable optimization and efficient inference with a small, fixed number of integration steps. Furthermore, by integrating open-vocabulary perception with geometry-consistent pose prediction and leveraging large-scale real and synthetic 6D datasets, OMNI-PoseX achieves scalable and real-time category-level pose estimation suitable for embodied robotic tasks.

\section{Methods}
\label{headings}

\begin{figure*}[t]
    \centering
    \includegraphics[width=1\linewidth]{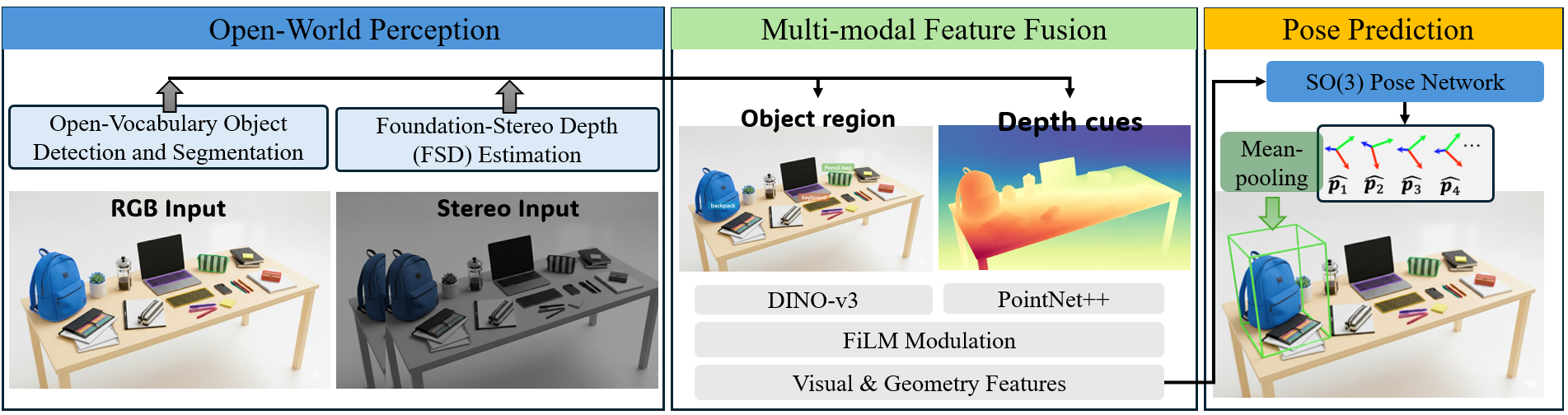}
    \caption{Network Architecture of the OMNI-PoseX.}
    \label{fig:framework}
\end{figure*}

OMNI-PoseX consists of two main components: an \textbf{open-world visual perception module} and an \textbf{SO(3)-aware pose prediction network}. Fig \ref{fig:framework} illustrates the overall framework of OMNI-PoseX.

The \textbf{open-world visual perception module} is responsible for object-level understanding and geometry extraction. Specifically, we adopt an open-vocabulary visual segmentation model together with a foundation stereo depth estimation network to generate object masks and dense depth maps. This module provides rich visual and geometric features, including object regions and depth cues, for downstream pose estimation. Since it is trained on large-scale public open-world datasets, it exhibits strong generalization ability to novel objects, categories, and environments.

The \textbf{SO(3) pose prediction network} takes the features extracted by the perception module as input and fuses multi-modal visual information to infer object pose in 3D space. It predicts the object’s orientation on the SO(3) manifold, as well as the 3D bounding box center and size. By decoupling open-world perception from geometry-aware pose estimation, OMNI-PoseX achieves robust and scalable 6D pose estimation under diverse and unconstrained settings.

\subsection{Lightweight Multi-modal Feature Fusion}
\label{sec:light_fusion}

To obtain an object-centric latent embedding suitable for downstream 
prediction tasks, we adopt a lightweight geometry-dominant fusion strategy. 
The design emphasizes the preservation of 3D structural properties while 
incorporating semantic priors in a computationally efficient manner. 
Rather than relying on heavy cross-attention or dual-stream architectures, 
our method conditions a pure geometric representation on compact semantic 
information, enabling both rotation-sensitive feature encoding and fast inference.

\textbf{Semantic Encoding:} 
Given an RGB image, we extract patch-level features $F \in \mathbb{R}^{H' \times W' \times D}$ using a frozen DINO encoder \cite{zhang2022dino}. A corresponding object mask $M$ identifies the target region, over which we perform mask-guided pooling and compress the pooled features via a lightweight projection:
\begin{equation}
\mathbf{s}' = \phi_s \Big( \frac{1}{|M|} \sum_{i \in M} F_i \Big), 
\quad \mathbf{s}' \in \mathbb{R}^{d_s}.
\end{equation}
This semantic vector captures category-level and contextual cues without 
introducing spatial dependencies, serving as a compact conditioning signal.

\textbf{Geometry Encoding:} 
The 3D points of the object, $P \subset \mathbb{R}^3$, are first sampled 
using farthest point sampling to obtain $P_{512}$. 
These points are encoded with PointNet++ to extract a rotation-sensitive 
geometric feature:
\begin{equation}
\mathbf{g} = \phi_g(P_{512}), \quad \mathbf{g} \in \mathbb{R}^{d_g}.
\end{equation}
By maintaining a purely geometric stream, the network preserves the 
structural information critical for SO(3)-aware reasoning and downstream 
pose or flow prediction.

\textbf{Feature Modulation and Latent Embedding:} 
We condition the geometric feature $\mathbf{g}$ on the semantic vector $\mathbf{s}'$ 
via a Feature-wise Linear Modulation (FiLM):
\begin{equation}
\mathbf{z}_{\text{obj}} = \phi_z \big( \psi(\mathbf{s}') \odot \mathbf{g} + \psi'(\mathbf{s}') \big),
\quad \mathbf{z}_{\text{obj}} \in \mathbb{R}^{d_z}.
\end{equation}
Here, $\psi$ and $\psi'$ are lightweight projections generating scaling and shift parameters, respectively. The resulting latent embedding $\mathbf{z}_{\text{obj}}$ fuses semantic and geometric cues in a lightweight, linear-complexity  manner while preserving rotation-equivariant structure for downstream SO(3)-aware pose estimation tasks.

\subsection{SO(3)-Aware Reflected Flow Matching}
This section introduce the \emph{SO(3)-aware Reflected Flow Matching} framework that learns rotation dynamics directly in the Lie algebra, ensuring geodesic consistency and stable training.

\paragraph{Geodesic Rotation Path on SO(3)}
Given an initial rotation $R_0 \in \mathrm{SO}(3)$ and a target rotation $R_1 \in \mathrm{SO}(3)$, we first define the relative rotation $\Delta R = R_0^{-1} R_1$, and map it to the Lie algebra using the logarithmic map $\omega = \log(\Delta R) \in \mathbb{R}^3$.
This representation corresponds to the minimal axis--angle parameterization of the relative rotation and defines a unique geodesic on $\mathrm{SO}(3)$.
To construct a time-symmetric trajectory, we introduce a reflected flow path with reflection time $t_r = 0.5$. The intermediate rotation $R(t)$ is defined as
\begin{equation}
R(t) =
\begin{cases}
R_0 \exp\!\left(2t\,\omega\right), & t \le 0.5, \\[4pt]
R_1 \exp\!\left(2(1-t)(-\omega)\right), & t > 0.5 .
\end{cases}
\end{equation}
This formulation ensures that rotations evolve strictly along the geodesic connecting $R_0$ and $R_1$, while maintaining exact temporal symmetry around the midpoint.

\begin{figure*}[ht]
    \centering
    \includegraphics[width=1\linewidth]{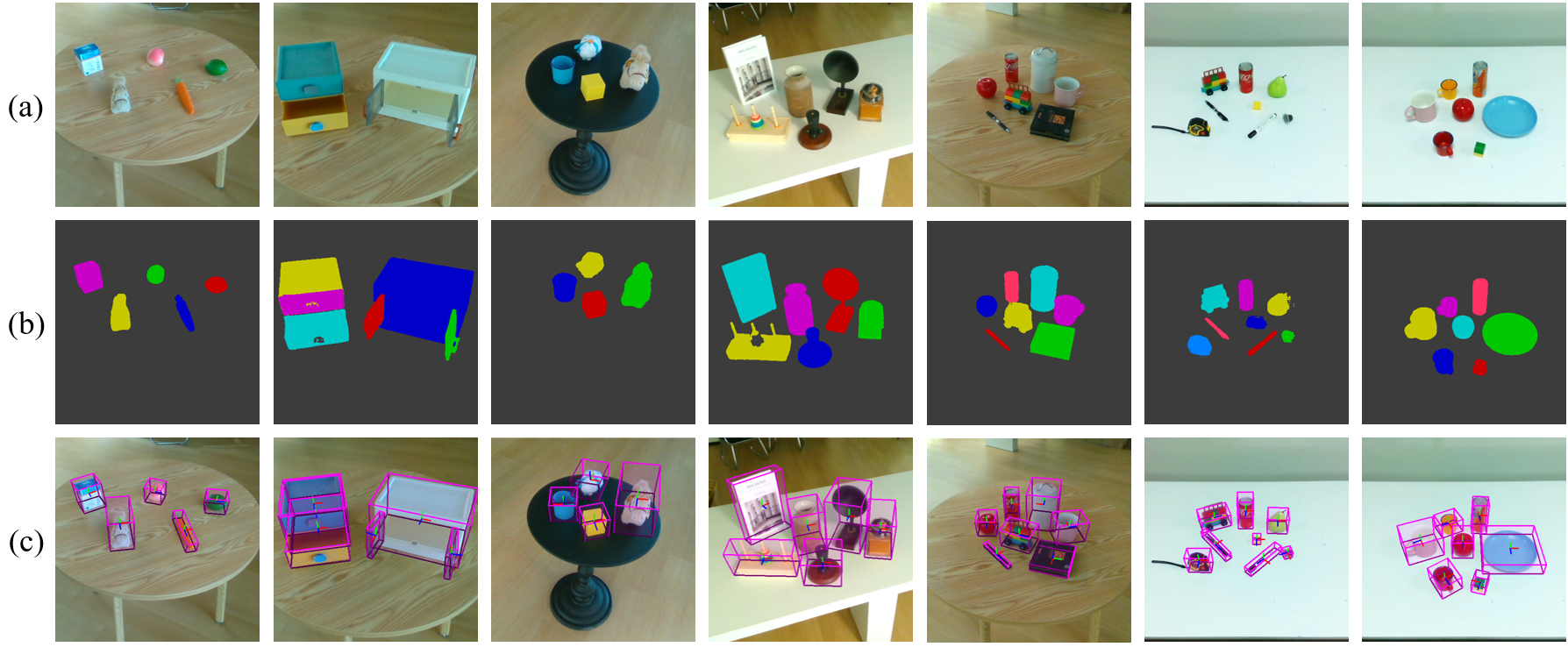}
    \caption{The figures illustrate the results of our model processing. (a) is the original RGB image. (b) is the mask segmentation result. (c) is the OMNI-PoseX prediction result.}
    \label{fig:compared}
\end{figure*}

\begin{table*}[ht]
\centering
\begin{threeparttable}

\caption{Performance evaluation on OMNI-6DPose benchmark}
\label{tab:omni6dpose_eval}
\small
\begin{tabular}{lcccccccc}
\toprule
\multirow{2}{*}{Variant}
& \multicolumn{3}{c}{AUC $\uparrow$}
& \multicolumn{2}{c}{VUS $\uparrow$}
& \multirow{2}{*}{Deg $\downarrow$}
& \multirow{2}{*}{Sht $\downarrow$}
& \multirow{2}{*}{time(ms)$^{*}$ $\downarrow$}\\
\cmidrule(lr){2-4} \cmidrule(lr){5-6}
& @25 & @50 & @75 & 5°/2cm & 5°/5cm & & \\
\midrule

NOCS ~\cite{wang2019normalized}      & 0.0 & 0.0 & 0.0 & 0.0 & 0.0 & - & - & -\\
SGPA ~\cite{chen2021sgpa}            & 10.5 & 2.0 & 0.0 & 4.3 & 6.7 & - & - & -\\
IST-Net ~\cite{liu2023net}         & 28.7 & 10.6 & 0.5 & 2.0 & 3.4 & - & - & -\\
HS-Pose ~\cite{zheng2023hs}         & 31.6 & 13.6 & 1.1 & 3.5 & 5.3 & - & - & -\\
GenPose++$^{*}$ (w/o E) \cite{zhang2024omni6dpose}  & 31.6 & 12.7 & 0.0 & 7.8 & 12.29 & 34.66 & 1.44 & 112 \\
GenPose++$^{*}$        & 40.37 & 20.46 & 2.2 & 10.33 & 15.29 & 32.11 & 1.33 & 178\\
\textbf{OMNI-PoseX (w/o E) $^{*}$ } 
                 & \textbf{35.6} & \textbf{15.43} & \textbf{2.17} 
                 & \textbf{11.34} & \textbf{13.83} 
                 & \textbf{33.36} & \textbf{1.43} & \textbf{17}\\
\textbf{OMNI-PoseX $^{*}$ } 
                 & \textbf{41.6} & \textbf{20.17} & \textbf{2.33} 
                 & \textbf{11.57} & \textbf{15.83} 
                 & \textbf{32.17} & \textbf{1.27} & \textbf{86}\\
\bottomrule
\end{tabular}

\begin{tablenotes}
\footnotesize
\item[$^{*}$] Sequences 0004 and 0302 in OMNI6D-pose are excluded due to data corruption.
\end{tablenotes}
\end{threeparttable}
\end{table*}

\paragraph{SO(3)-Aware Velocity Field}
Rather than defining velocities in the ambient matrix space, we express the flow velocity in the body-frame Lie algebra, yielding a constant velocity field:
\begin{equation}
v(t) =
\begin{cases}
+2\omega, & t \le 0.5, \\[4pt]
-2\omega, & t > 0.5 .
\end{cases}
\end{equation}
The velocity is independent of the current rotation state $R(t)$ and remains bounded across time, which significantly stabilizes optimization and avoids stiff ODE dynamics commonly observed in score-based generative formulations.

\paragraph{Flow Matching Objective on SO(3)}
Our network is trained to predict the instantaneous rotational velocity in the Lie algebra:
\begin{equation}
f_\theta(R_t, t) \;\rightarrow\; \hat{\omega}_t \in \mathbb{R}^3 .
\end{equation}
The training objective minimizes the squared error between the predicted and ground-truth velocities:
\begin{equation}
\mathcal{L}_{\mathrm{rot}}
=
\mathbb{E}_{t}
\left[
\left\|
\hat{\omega}_t -
\begin{cases}
+2\omega, & t \le 0.5, \\[4pt]
-2\omega, & t > 0.5
\end{cases}
\right\|^2
\right] .
\end{equation}

By supervising velocities in the Lie algebra rather than absolute rotations, the model learns a geometry-consistent vector field that naturally generalizes across object categories and orientations.

\subsection{Inference by RK2 Integration on $\mathrm{SO}(3)$}
At inference time, the learned SO(3)-aware velocity field is used to recover object orientation by integrating a first-order ordinary differential equation on the rotation manifold. Starting from an initial rotation $R(0) = R_0$, typically set to the identity rotation or a coarse prior, we evolve the rotation according to
\begin{equation}
\frac{dR(t)}{dt} = R(t)\,\widehat{f_\theta(R(t), t)},
\end{equation}
where $\widehat{(\cdot)}$ denotes the skew-symmetric matrix associated with a Lie algebra vector in $\mathbb{R}^3$.
The predicted angular velocity lies in the Lie algebra $so(3)$ and remains bounded over time, resulting in a non-stiff ODE on $SO(3)$. We adopt a second-order \textbf{Runge--Kutta (RK2 / Heun)} scheme to integrate the rotation flow efficiently while preserving geometric structure. At each step, the angular velocity is first evaluated at the current rotation to compute a provisional rotation. A second evaluation is then performed at this provisional state, and the final rotation update is obtained via the average of the two velocities applied through the matrix exponential. Formally, the discrete RK2 (Heun) update is:
\begin{equation}
\begin{aligned}
R_{k+1} &= R_k \exp\!\left( \frac{\Delta t}{2} (\hat{\omega}_k + \hat{\omega}_{k+1}) \right), \\
\hat{\omega}_k &= f_\theta(R_k, t_k), \\
\hat{\omega}_{k+1} &= f_\theta\!\left(R_k \exp(\Delta t \hat{\omega}_k), t_{k+1}\right).
\end{aligned}
\end{equation}
Due to the smooth and bounded velocity field, accurate rotation integration is achieved with only a few RK2 steps (we use 5), enabling stable and real-time 6D pose estimation in embodied tasks.

\subsection{Training on Large-Scale 6D-Pose Datasets}
OMNI-PoseX is trained on two complementary large-scale datasets, \textbf{Omni6DPose} and \textbf{Omni6D}, providing extensive object coverage, instance diversity, and material variation in both real world and simulated scenarios.
\textbf{Omni6DPose} includes \textbf{ROPE} (Real 6D Object Pose Estimation) with 332K RGB images and 1.5M pose annotations across 581 instances in 149 categories, capturing rich appearance and viewpoint variation, and \textbf{SOPE} (Simulated 6D Object Pose Estimation) with 475K synthetic images and 5M annotations over 4,162 instances, enhancing instance diversity.
\textbf{Omni6D} is a comprehensive RGB-D dataset covering 166 categories and 4,688 instances with canonical poses, 0.8M captures in cluttered scenes, improving robustness to environmental variation and supporting open-world generalization.

\section{Experiments}
We evaluate OMNI-PoseX on large-scale 6D pose datasets, unseen objects for zero-shot generalization, and long-horizon embodied tasks to demonstrate its open-world capability, real-time performance. Fig \ref{fig:compared}. shows the prediction results of OMNI-PoseX in real-world.

\begin{table*}[t]
\centering
\caption{Ablation Study on OMNI-PoseX.}
\label{tab:ablation_all}
\small
\begin{tabular}{llcccccc}
\toprule
\multirow{2}{*}{Category} & \multirow{2}{*}{Variant}
& \multicolumn{3}{c}{AUC $\uparrow$}
& \multicolumn{2}{c}{VUS $\uparrow$} \\
\cmidrule(lr){3-5} \cmidrule(lr){6-7}
& & @25 & @50 & @75 & 5°/2cm & 5°/5cm & \\
\midrule

\multirow{3}{*}{Rotation}
& Continuous representation     & 39.2 & 17.25 & 1.89 & 9.88 & 14.23 \\
& \textbf{SO(3) representation} & 41.6 & 20.17 & 2.33 & 11.57 & 15.83 \\
\midrule

\multirow{4}{*}{Fusion}
& DINO only                 & 11.3 & 2.5 & 0.0 & 1.4 & 2.1 \\
& PointNet++ only           & 37.3 & 15.3 & 1.7 & 0.0 & 0.0 \\
& Pointwise Fusion          & 41.6 & 20.08 & 2.16 & 10.98 & 15.22 \\
& \textbf{FiLM Fusion}      & 41.6 & 20.17 & 2.33 & 11.57 & 15.83 \\
\midrule

\multirow{2}{*}{Depth}
& Real Depth                    & 41.6 & 20.17 & 2.33 & 11.57 & 15.83 \\
& Foundation-Stereo Depth       & 41.57 & 20.21 & 2.45 & 11.60 & 15.85 \\
\midrule

\multirow{2}{*}{Data Scale}
& Omni6DPose                    & 41.6 & 20.17 & 2.33 & 11.57 & 15.83 \\
& \textbf{Omni6DPose + Omni6D} 
                                & \textbf{42.2} & \textbf{20.2} & \textbf{2.31} 
                                & \textbf{11.63} & \textbf{15.88} \\
\bottomrule
\end{tabular}
\end{table*}

\subsection{Evaluation on 6D Pose Estimation Benchmark}

Following the OMNI-6DPose benchmark, we report AUC under ADD-S thresholds (0.25, 0.5, 0.75) and VUS under combined rotation–translation criteria (5°2cm, 5°5cm).
Table~\ref{tab:omni6dpose_eval} reports quantitative results on the OMNI-6DPose benchmark. We compare OMNI-PoseX with representative category-level pose estimation methods, including NOCS, SGPA, IST-Net, HS-Pose, and GenPose++. All methods are trained on SOPE and directly evaluated on ROPE to assess cross-domain generalization.

Compared with prior methods, GenPose++ achieves strong performance across all AUC and VUS metrics, confirming the advantage of distribution modeling for category-level pose estimation. By modeling pose as a generative process, it better captures symmetry-induced ambiguity and reduces reliance on depth-based fitting.
OMNI-PoseX further improves performance while significantly reducing inference cost. Without external enhancement (w/o E), OMNI-PoseX achieves higher VUS@5°/2cm (11.34) than GenPose++ (10.33) with substantially lower inference time (17 ms vs. 178 ms). With enhancement enabled, OMNI-PoseX attains the best overall AUC@25 (41.6) and competitive AUC@50 (20.17), while maintaining lower rotation error (32.17°) and translation shift (1.27 cm). Notably, inference time is reduced to 86 ms, which is approximately half of GenPose++ under comparable settings.
These results indicate that learning a geometry-consistent velocity field on $SO(3)$ enables accurate pose prediction with a small, fixed number of integration steps, preserving the robustness advantages of generative modeling while substantially improving computational efficiency.

\noindent\textbf{Time effciency:} Compared with GenPose++, OMNI-PoseX achieves higher inference efficiency. Without energynet, OMNI-PoseX reduces runtime from 112 ms to 17 ms while improving AUC and VUS metrics; with energynet deployed, it attains comparable or better accuracy than GenPose++ (86 ms vs. 178 ms) at roughly half the inference time.

\subsection{Ablation Study}
To systematically evaluate the contribution of each core component in OMNI-PoseX, we conduct a unified ablation study covering three aspects: (1) rotation modeling strategy, (2) multi-modal feature fusion design, and (3) training data composition. Results are summarized in Table~\ref{tab:ablation_all}.

Table~\ref{tab:ablation_all} reports ablation results on rotation modeling, fusion strategy, depth source, and data scale. Replacing a continuous rotation parameterization with the proposed $SO(3)$ representation consistently improves all metrics (AUC@25: 39.2 $\rightarrow$ 41.6; VUS@5$^\circ$/2cm: 9.88 $\rightarrow$ 11.57), with more pronounced gains under stricter thresholds. Notably, AUC in pose evaluation is more sensitive to translation accuracy and bounding box scale alignment, whereas VUS places stronger emphasis on angular precision. The larger improvements observed in VUS indicate that geometry-aware rotation modeling primarily enhances orientation estimation while maintaining stable translation performance.

For feature fusion, DINO-only performs poorly, suggesting that semantic cues alone are insufficient for precise geometric reasoning. PointNet++ achieves competitive AUC but degrades significantly under strict VUS metrics, indicating limitations in resolving rotational ambiguity. Multi-modal fusion substantially improves both AUC and VUS, demonstrating the complementary roles of semantic and geometric features. FiLM-based modulation further provides consistent gains over pointwise fusion, implying more effective cross-modal conditioning. In addition, foundation stereo depth achieves nearly identical performance to real depth, confirming robustness to depth quality. Expanding training data from Omni6DPose to Omni6DPose + Omni6D yields consistent improvements (AUC@25: 41.6 $\rightarrow$ 42.2), suggesting that the framework scales with increased instance diversity.

\begin{figure}[h]
    \centering
    \includegraphics[width=\linewidth]{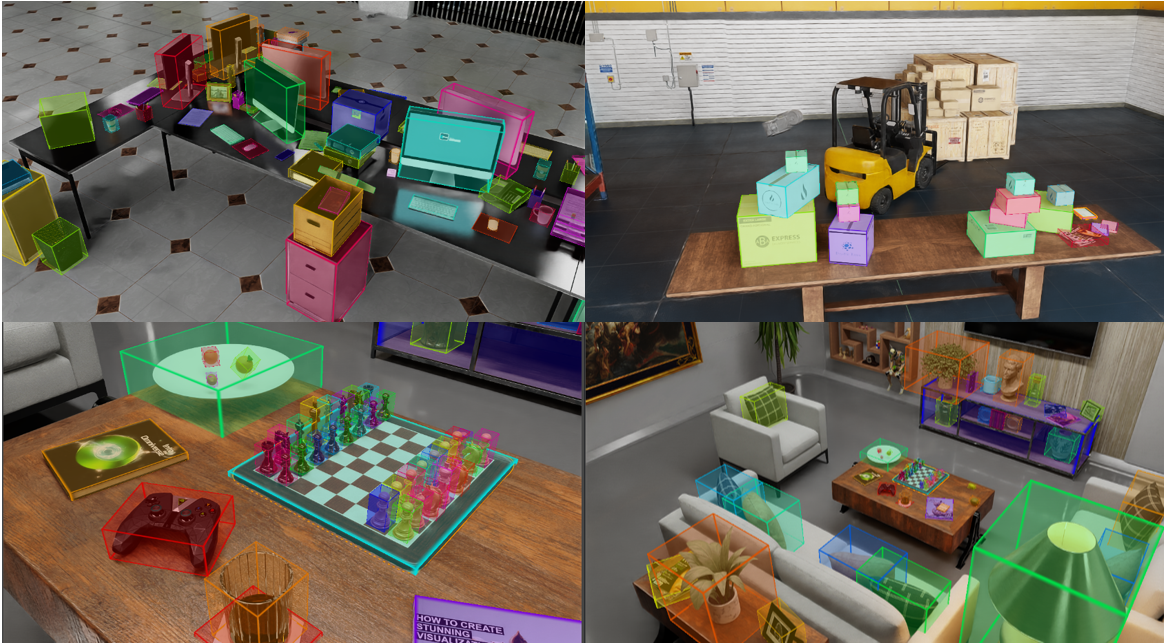}
    \caption{Samples of unseen objects in Issac-Sim.}
    \label{fig:isaac_sim}
\end{figure}

\begin{figure*}[t]
    \centering
    \includegraphics[width=1.0\textwidth, height=0.47\textheight]{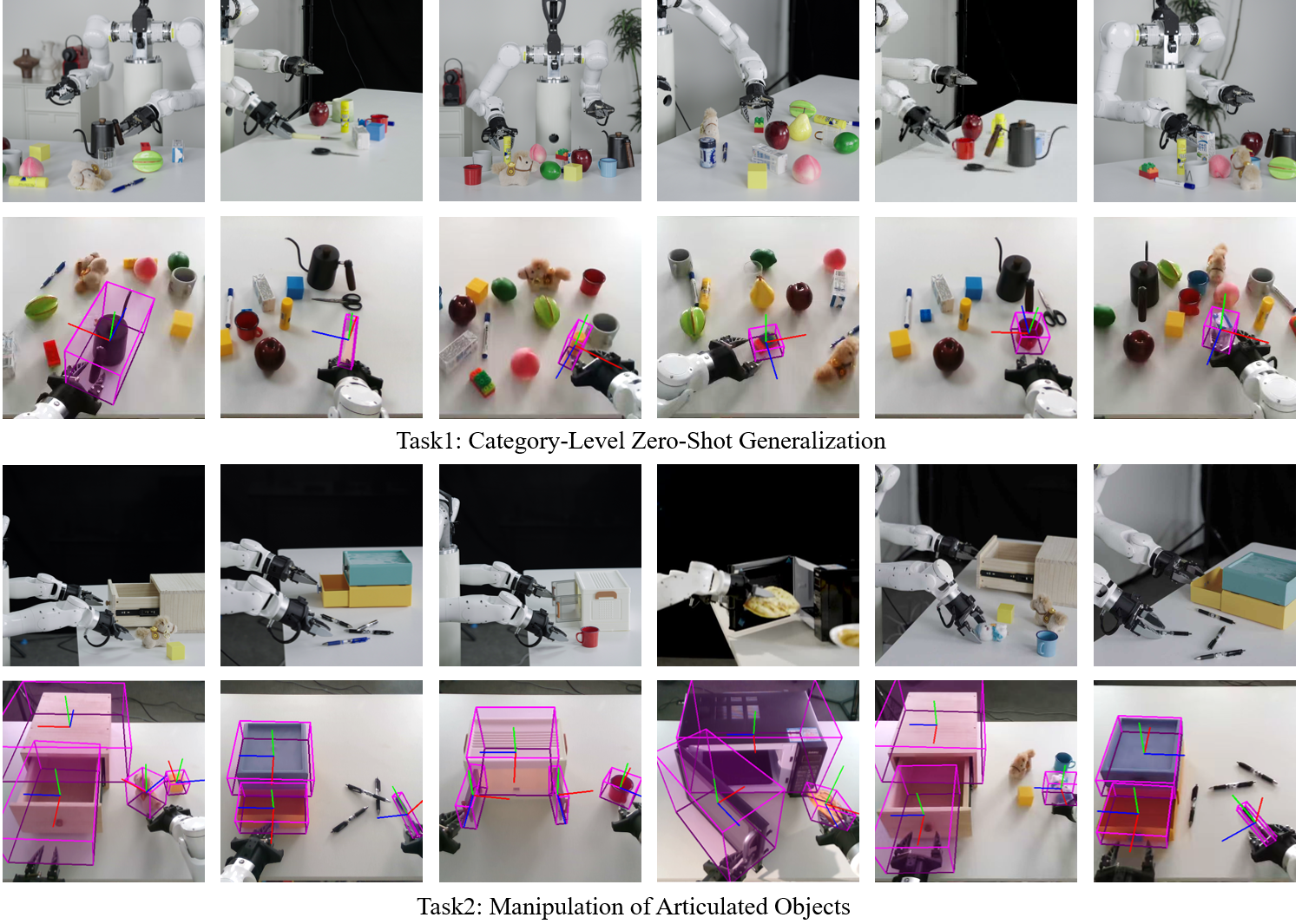}
    \caption{Real-world demonstrations of OMNI-PoseX in daily manipulation tasks. \textbf{Top}: Category-level zero-shot grasping across unseen object instances, validating cross-category generalization under geometric and appearance variations. \textbf{Bottom}: Articulated object manipulation (drawer and cabinet door), requiring stable 6D pose tracking to support constrained motion and contact-consistent execution.}
    \label{fig:Embodied}
\end{figure*}

\begin{table*}[t]
\centering
\caption{Zero-Shot Generalization on Unseen Embodied Tasks (Isaac Sim).}
\label{tab:zero_shot}
\small
\begin{tabular}{lccccccc c}
\toprule
\multirow{2}{*}{Variant}
& \multicolumn{3}{c}{AUC $\uparrow$}
& \multicolumn{2}{c}{VUS $\uparrow$}
& \multirow{2}{*}{Deg $\downarrow$}
& \multirow{2}{*}{Sht $\downarrow$}
& \multirow{2}{*}{Time (ms) $\downarrow$} \\
\cmidrule(lr){2-4} \cmidrule(lr){5-6}
& @25 & @50 & @75 & 5°/2cm & 5°/5cm & & & \\
\midrule
SGPA           & 10.1 & 1.9  & 0.1  & 2.58 & 5.32 & -  & - & - \\
IST-Net           & 22.4 & 6.9  & 0.5  & 3.85 & 7.32 & -  & - & - \\
HS-Pose            & 25.2 & 7.7  & 0.6  & 4.42 & 9.15 & -  & - & - \\
GenPose++       & 33.97 & 11.17 & 1.21 & 6.89 & 15.52 & 38.02 & 3.04 & 178 \\ 

\textbf{OMNI-PoseX (Ours)} 
                        & \textbf{35.6} 
                        & \textbf{12.17} 
                        & \textbf{1.33} 
                        & \textbf{7.57} 
                        & 15.46 
                        & \textbf{36.17} 
                        & \textbf{2.75} 
                        & \textbf{86} \\

\bottomrule
\end{tabular}
\end{table*}

\subsection{Zero-Shot Generalization on Unseen Datasets}
To evaluate the zero-shot generalization capability of OMNI-PoseX in open-world (without finetune) embodied settings, we construct a simulated 6D pose benchmark using Isaac Sim, as shown in Fig \ref{fig:isaac_sim}. 
The benchmark consists of 20 scenes spanning three distinct domains: supermarket, household, and logistics. Across these scenes, more than 30 object categories are included, all of which are unseen during training. The objects exhibit diverse geometries, materials, scales, and placements, with variations in lighting, occlusion, and scene layout.

Table~\ref{tab:zero_shot} demonstrates that OMNI-PoseX consistently outperforms GenPose++ in zero-shot embodied settings without any finetuning. The proposed method achieves higher AUC scores across all thresholds and improves VUS@5°/2cm (7.57 vs.~6.89), while reducing both rotational error (36.17$^\circ$ vs.~38.02$^\circ$) and translational shift (2.75\,cm vs.~3.04\,cm). These gains indicate enhanced robustness to domain shifts in object categories, scene layouts, and visual conditions. These results suggest that the geometry-consistent formulation of OMNI-PoseX enhances robustness particularly when transferring to entirely unseen datasets.

Failure cases are primarily associated with elongated or thin planar objects, such as cabinet doors or slender tools. For such geometries, small orientation deviations around the principal axis can produce large spatial displacements at object extremities, thereby amplifying rotational errors. Moreover, inaccurate size estimation for thin or flat structures may introduce bias in translation recovery, resulting in coupled errors between orientation prediction and scale-aware alignment. These findings suggest that long-axis ambiguity and thickness estimation remain challenging under severe domain variation, motivating future exploration of stronger geometric priors and scale-consistent constraints.

\subsection{Task-Level Evaluation on Embodied Manipulation Tasks}
We build an open-world manipulation pipeline, termed as \textbf{ANYGRASP}, built upon OMNI-PoseX. In this system, OMNI-PoseX serves as the core perception module, providing 6D pose estimates directly to the grasp planner. The entire pipeline operates in a zero-shot setting, without any task-dependent fine-tuning, enabling fully closed-loop manipulation, as shown in Fig \ref{fig:Embodied}.
To assess system-level robustness and generalization, we construct two categories of embodied manipulation tasks:

\textbf{Category-Level Zero-Shot Generalization}:  
The robot performs grasping on general object instances spanning diverse geometries, scales, and materials, evaluating cross-category transfer and open-world robustness. Example tasks include: \textit{"grasping a mug from a new set"}, \textit{"picking up a novel toy block"}, and \textit{"lifting an apple from the basket"}.


\textbf{Manipulation of Articulated Objects}:  
The robot interacts with articulated structures such as drawers and cabinet doors, requiring temporally stable and geometrically consistent pose estimation to support contact-rich manipulation and constrained motion execution. Example tasks include: \textit{"opening a drawer and picking an mug inside"} and \textit{"closing a cabinet door after placing items"}.

Experimental results demonstrate reliable and precise task completion across given settings, while qualitative rollouts illustrating representative successes are provided in the supplementary video.

\section{Conclusion}
We present \textbf{OMNI-PoseX}, a vision foundation model for open-world 6D pose estimation that incorporates rotational geometry via an SO(3)-aware reflected flow formulation. By learning velocity fields in the Lie algebra and leveraging geometry-dominant multi-modal fusion with large-scale supervision, OMNI-PoseX enables stable, efficient, and geometry-consistent pose inference in diverse open-world settings.
Extensive experiments demonstrate that OMNI-PoseX achieves state-of-the-art performance on large-scale 6D pose benchmarks, while maintaining efficient inference with a small, fixed number of integration steps. Beyond benchmark evaluation, we further validate its system-level effectiveness through integration into an open-world manipulation framework (ANYGRASP). Across category-level zero-shot grasping, cluttered scene manipulation, and articulated object interaction, OMNI-PoseX provides temporally stable and geometrically consistent pose predictions that enable reliable closed-loop robotic execution. These results highlight the practical value of geometry-aware pose foundation models for scalable embodied manipulation.

\textbf{Limitations of current work:} Although OMNI-PoseX demonstrates strong open-world performance, it still inherits limitations from current open-vocabulary visual backbones. Segmentation quality and semantic generalization may degrade under severe domain shifts, heavy occlusion, cluttered scenes, or visually ambiguous categories with subtle inter-class differences. Because pose estimation depends on accurate object localization and feature consistency, perception errors can propagate to downstream geometric reasoning, affecting both orientation and translation accuracy. In addition, the proposed lightweight fusion design prioritizes computational efficiency and structural preservation, but does not explicitly model higher-order cross-modal interactions between visual semantics and geometric features. The current fusion strategy mainly performs feature-level aggregation without enforcing geometry-aware alignment or uncertainty modeling across modalities, which may limit robustness in challenging scenarios involving partial visibility, symmetry ambiguity, or scale inconsistency.

\textbf{Future works:} Future work will explore more expressive and geometry-aware fusion mechanisms, such as manifold-consistent cross-attention, uncertainty-aware multi-modal reasoning, or iterative refinement schemes that jointly optimize perception and pose estimation. Improving the robustness and calibration of open-world perception models under large distribution shifts will also be critical for enabling more reliable deployment in diverse embodied environments.


\bibliographystyle{IEEEtran}
\bibliography{ref}

\end{document}